\documentclass[10pt,twocolumn,letterpaper]{article}

\usepackage{cvpr}
\usepackage{times}
\usepackage{epsfig}
\usepackage{graphicx}
\usepackage{amsmath}
\usepackage{amssymb}

\usepackage{array}
\usepackage{multirow}
\usepackage{mmstyles}
\usepackage{xcolor}
\usepackage{url}  

\DeclareMathAlphabet{\mathpzc}{OT1}{pzc}{m}{it} 

\newcolumntype{P}[1]{>{\centering\arraybackslash}p{#1}}
\newcolumntype{M}[1]{>{\centering\arraybackslash}m{#1}}

\makeatletter
\def\hlinew#1{%
  \noalign{\ifnum0=`}\fi\hrule \@height #1 \futurelet
   \reserved@a\@xhline}
\makeatother

\usepackage[breaklinks=true,bookmarks=false]{hyperref}

\cvprfinalcopy 

\def\cvprPaperID{1800} 

\begin{document}

\title{Context and Attribute Grounded Dense Captioning}

\author{ Guojun Yin$^{1}$, Lu Sheng$^{2, 4}$, Bin Liu$^1$\thanks{Bin Liu is the corresponding author.} , Nenghai Yu$^1$, Xiaogang Wang$^2$, Jing Shao$^3$ \\
\normalsize $^1$University of Science and Technology of China, Key Laboratory of  Electromagnetic Space Information, \\ \normalsize The Chinese Academy of Sciences, $^2$CUHK-SenseTime Joint Lab, The Chinese University of Hong Kong, \\
\normalsize $^3$SenseTime Research, $^4$College of Software, Beihang University \\
\tt\small gjyin@mail.ustc.edu.cn, lsheng@buaa.edu.cn, \\ 
\tt\small \{flowice,ynh\}@ustc.edu.cn, xgwang@ee.cuhk.edu.hk, shaojing@sensetime.com
}

\maketitle

\begin{abstract}

Dense captioning aims at simultaneously localizing semantic regions and describing these regions-of-interest (ROIs) with short phrases or sentences in natural language.
Previous studies have shown remarkable progresses, but they are often vulnerable to the \emph{aperture} problem that a caption generated by the features inside one ROI lacks \emph{contextual coherence} with its surrounding context in the input image.
In this work, we investigate contextual reasoning based on multi-scale message propagations from the neighboring contents to the target ROIs.
To this end, we design a novel end-to-end context and attribute grounded dense captioning framework consisting of 1) a contextual visual mining module and 2) a multi-level attribute grounded description generation module.
Knowing that captions often co-occur with the linguistic attributes (such as \emph{who}, \emph{what} and \emph{where}), we also incorporate an auxiliary supervision from hierarchical linguistic attributes to augment the distinctiveness of the learned captions.
Extensive experiments and ablation studies on Visual Genome dataset demonstrate the superiority of the proposed model in comparison to state-of-the-art methods.

\end{abstract}

\section{Introduction}
\label{sec:intro}

\begin{figure}[t]
\centering
\includegraphics[width=0.85\linewidth]{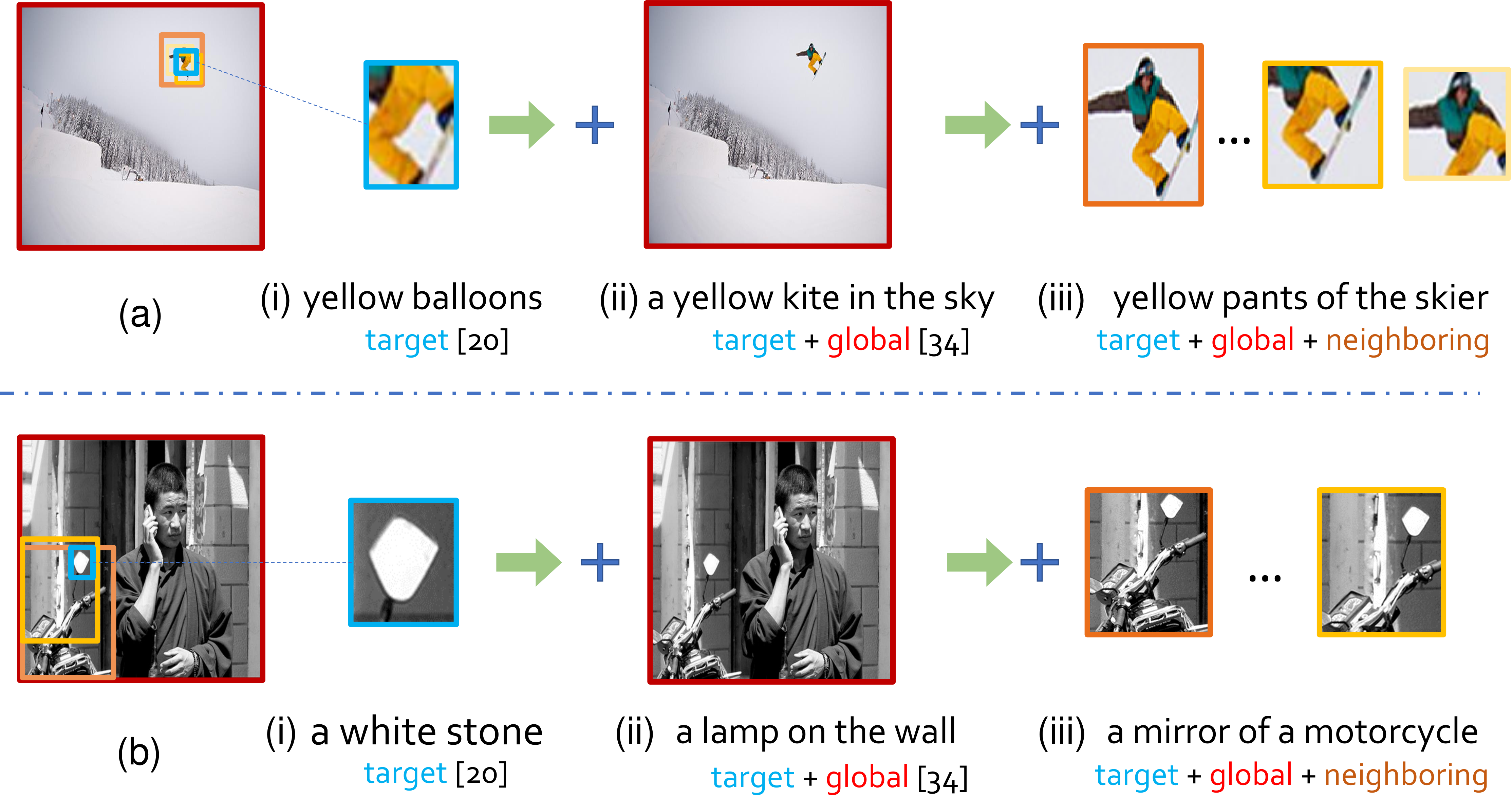}
\caption{Dense captioning with different levels of contextual interactions: (i) without any contextual cues (marked by {\color{cyan} blue})~\cite{johnson2016densecap}, (ii) with guidance from the global cue (marked by {\color{red} red})~\cite{yang2017densecap}, and (iii) with mutual interactions from neighboring (marked by \textcolor{orange} {orange}) and global visual information. (Best viewed in color.)
}
\label{fig:fig1}
\end{figure}

Dense captioning, which was first introduced by \cite{johnson2016densecap}, is to parse semantic contents in an input image and describe them with captions in natural languages.
It can benefit other tasks, including image captioning~\cite{you2016semanticattention}, segmentation~\cite{mottaghi2014role}, visual question answering~\cite{Goyal_2017_CVPR} and \textit{etc}.
In this paper, we mainly focus on the caption generation and adopt Faster RCNN~\cite{ren2015faster} for semantic instances localization, following recent advances~\cite{johnson2016densecap,yang2017densecap}.

Differing from subjective image descriptions for high-level abstraction of an entire image, captions of semantic instances in compact bounding boxes are far more objective and less affected by ambiguities raised by subjective annotations.
That is, incorrect captions may be generated when the target regions are visually ambiguous without contextual reasoning. 
For example, it may falsely caption the target ROIs marked in blue-box as ``yellow balloons'' rather than ``yellow pants'' in Fig.~\ref{fig:fig1}(a-i), if not aware of their contextual visual contents~\cite{johnson2016densecap}.
An alternative solution proposed in~\cite{yang2017densecap} try to exploit the global feature from the entire image as the contextual cue to improve the region captioning. However, the descriptions sometimes are corrupted by global appearance, especially for small and unusual objects against dominant global contents. The ``yellow pants'' in Fig.~\ref{fig:fig1}~(a-ii) is mistakenly described as ``yellow kite in the sky''. The similar phenomenon happens in Fig.~\ref{fig:fig1}~(a-ii) that it mistakenly generates ``a mirror'' rather than ``a lamp''.

In contrast to the prior arts, in this study, we show that the innovative model,
named as \textit{Context and Attribute Grounded Network} (CAG-Net),
designed with contextual correlated visual cues (\ie, \emph{local, neighboring, global}) permits multi-scale contextual message passing to reinforce regional description generation.
For example, the neighboring ROIs marked in warm-box in Fig.~\ref{fig:fig1}(a-iii), semantically connecting to the visual features in the target in blue-box in Fig.~\ref{fig:fig1}(a-i), provide more valuable hints that the target is a ``yellow pants'' belonging to a skier.
Such contextual learning has shown its remarkable potential in other tasks including object detection, segmentation and retrieval. However, the learning of contextual representation, and how it can effectively function on dense captioning, remains an open problem.
Specifically, the proposed CAG-Net consists of two vital modules:

1) \textit{Contextual Feature Extractor}, establishing a non-local similarity graph for the feature interaction between the target ROI and its neighboring ROIs based on their feature affinity and spatial nearness, allows adaptive contextual information sharing from multiple adjacent ROIs (\ie, global and neighbors) to interact with the target ROI.

2) \textit{Attribute-Grounded Caption Generator} adopts LSTM as the fundamental unit and fuses contextual features to generate the description for the target ROI. To reinforce the coarse-to-fine structure of description generation, we adopt coarse-level and fined-level linguistic attribute losses as the additional supervision respectively at the sequential LSTM cells.
Without sequential restrictions from the ground-truth captions, such keywords or attributes are more recognizable by the content in the target ROI, and thus own a more stable discriminative power for the extraction of visual patterns. To some extent, it is similar with the visual attributes of objects in multi-label classification.

Our contributions are listed as follows:

%
\noindent 1) We design a context and attribute grounded dense captioning model that permits multi-scale (\ie, local, neighboring, global) contextual information sharing and message passing, in which the knowledge integration is built on a non-local similarity graph among instances in the input image.
%

\noindent 2) A coarse-to-fine linguistic attribute supervision is proposed to enhance the discriminativeness of the generated captions, in which the ground-truth hierarchical linguistic attributes are matched to the predicted keywords through a
novel coarse-to-fine manner.
%

\noindent 3) Extensive experiments demonstrate the effectiveness of the proposed CAG-Net model on the challenging large-scale VG dataset.

\section{Related Work}
\label{sec:related_work}

Image captioning to describe a general image with natural language was explored in recent years~\cite{Chen_2017_CVPR,Lu_2017_CVPR,Gan_2017_CVPR,Ren_2017_CVPR,Aneja_2018_CVPR,Luo_2018_CVPR,Yao_2018_ECCV}.
The works~\cite{Chen_2017_CVPR,yao2016boostingqattribute,Anderson_2018_CVPR,Chen_2018_ECCV,Jiang_2018_ECCV,Chen_2018_ECCV1} focused on improve ifiguremage captioning by the attention-embedded features generated by an additional attention model. Based on the attention model, Gu~\etal~\cite{gu2017stack} adopted a coarse-to-fine framework which increased the model complexity gradually with increasingly refined attention weights for image captioning. 
In our work, dense image captioning renders individual captions for different ROIs in the image.
As for dense captioning, we firstly adopt the multi-scale feature interaction and attribute grounded generation for accurate region descriptions.
Our coarse-to-fine strategy is based on the hierarchical attribute supervisions rather than the different attention inputs of the description generation modules~\cite{gu2017stack}.
The previous works~\cite{you2016semanticattention,yao2016boostingqattribute} adopted the attributes (the words in the vocabulary) to train an additional model for another input of the LSTM cells for the description generation. 
Differing from that, our work adopts the linguistic attributes as the auxiliary supervision for coarse-to-fine generation without any external branches or input for captioning.

\noindent \textbf{Dense Image Captioning.}
Dense image captioning is supposed to not only localize the regions of interest in the image but also generate descriptions with natural language, which was first proposed in~\cite{johnson2016densecap}.
Johnson~\etal~\cite{johnson2016densecap} introduced a new dense localization layer, which was fully differentiable and used bilinear interpolation to smoothly extract the activations inside each region.
Yang~\etal~\cite{yang2017densecap} exploited more accurate localization for regions by joint inference of localization and description for a given region proposal, while the global feature of the image was used as the contextual cues to improve region captioning.
However, these previous works did not capture the relative features of different regions and valid message passing between contextual regions for accurate region captioning.

\noindent \textbf{Contextual Learning.}
Contextual learning was employed in various topics in recent years~\cite{tang2017multi,liu2018picanet,li2016human,Zhang_2018_CVPR,Wei_2018_ECCV,Dvornik_2018_ECCV,desai2011discriminative},~\eg,~object detection, segmentation, and retrieval.
For instance, the visual features captured from a bank of object detectors are combined with global features by~\cite{li2010object}.
For both detection and segmentation, learning feature representations from a global view rather than the located object itself has been proven effective by~\cite{gjyin_eccv2018,mottaghi2014role}.
Gkioxari~\etal~\cite{gkioxari2015contextual} used more than one region proposals for action recognition  while Hu~\etal~\cite{hu2018relation} processed a set of objects simultaneously through interaction between their appearance feature and geometry, thus allowing modeling of their relations for object detection.
As for contextual learning for image captioning, Yao \etal ~\cite{Yao_2018_ECCV} computed the probability distribution on all the semantic relation classes for each object pair with the learnt visual relation classifier to establish the semantic graph for image captioning. 
Contextual feature learning among the located regions for dense captioning has never been explored in the previous works.
In our work, we establish the contextual message passing module without any additional branches or any auxiliary relation labels.

\begin{figure*}[t]
\centering
\includegraphics[width=1.0\linewidth]{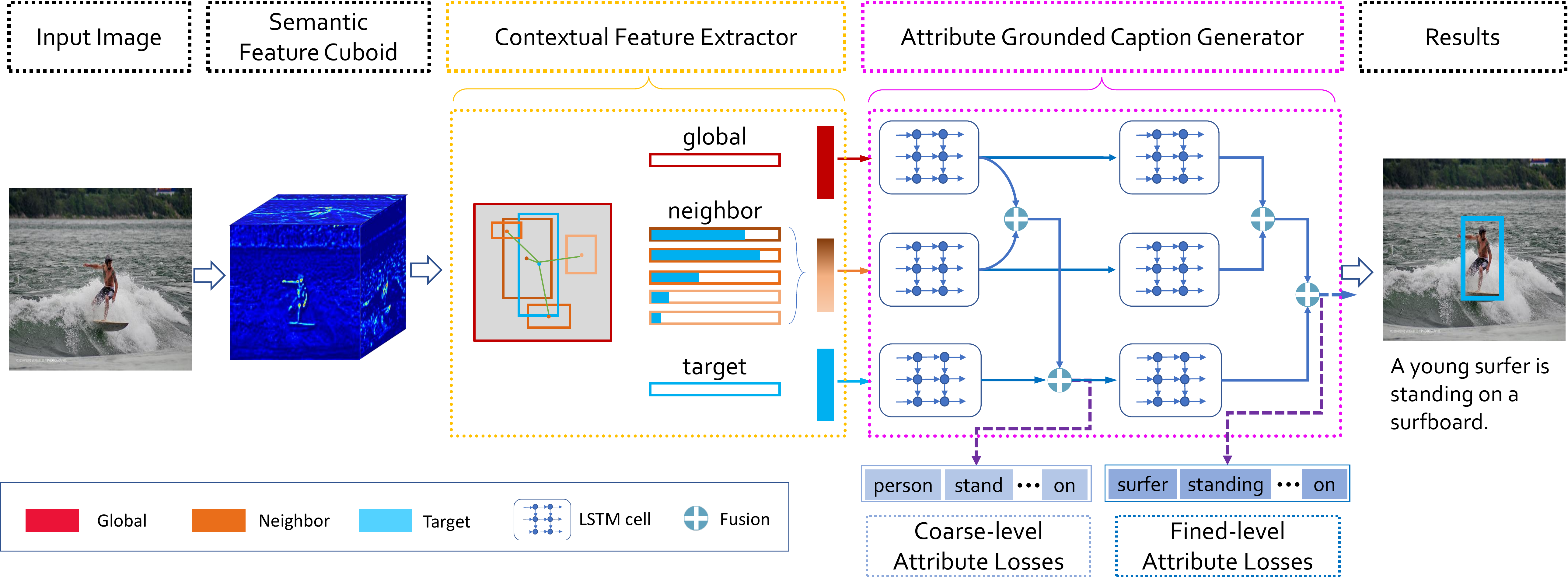}
\caption{The architecture of CAG-Net. The multi-scale features are generated by the proposed Contextual Feature Extractor after region proposals. Then the \textcolor{cyan}{\textit{local}} (in blue) feature of the target region and multi-scale context cues, \ie, \textcolor{red}{\textit{global}} (in red) and \textcolor{orange}{\textit{neighboring}} (in orange), broadcast into the Attribute Grounded Caption Generator for region captioning in parallel. The final descriptions of the target region are generated jointly by the hierarchical structures trained with the auxiliary attribute losses.
}
\label{fig:pipeline}
\end{figure*}

\section{Context and Attribute Grounded Dense Captioning (CAG-Net)}
\label{sec:pipeline}

\begin{figure}[t]
\centering
\includegraphics[width=1.0\linewidth]{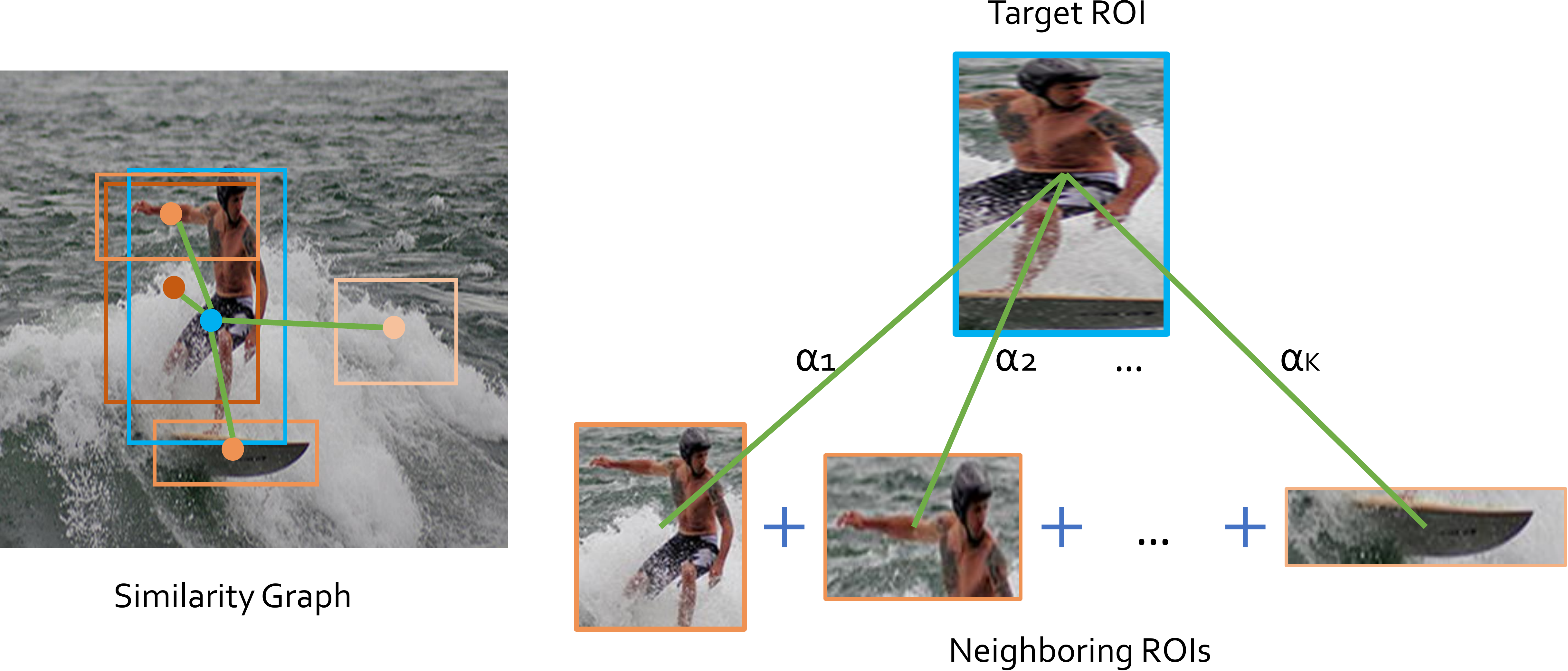}
\caption{
An example of Contextual Feature Extractor for the {target proposal}. (left) The similarity graph between {\color{cyan}target} (in blue) proposal  and contextual {\color{orange}neighboring} (in orange) proposals are generated considering both spatial configuration and appearance similarity. (right) The \textit{neighboring} feature are obtained by fusing the {contextual neighboring proposals} with the similarity graph. Best viewed in color.}
\label{fig:context_mining}
\end{figure}

In this paper, we propose a novel end-to-end dense image captioning framework, named as \textit{Context and Attribute Grounded Dense Captioning} (CAG-Net).
As shown in Fig.~\ref{fig:pipeline}, we first learn visual features of the input image by a CNN model as the way adopted by Faster RCNN~\cite{ren2015faster}, and obtain the semantic features. Such semantic features are used to generate a set of candidate regions (ROIs) by a Region Proposal Network (RPN)~\cite{ren2015faster}. 
Based on these ROI features, we introduce a \textit{Contextual Feature Extractor } (CFE) which generates the global, neighboring and local (\ie, target itself) cues (Sec.~\ref{subsec:context}). The neighboring cues are collected by establishing a similarity graph between the target ROI and the neighboring ROIs, shown in Fig.~\ref{fig:context_mining}.
The multi-scale contextual cues, broadcast in parallel, are fused by an \textit{Attribute Grounded Caption Generator} (AGCG) which employs multiple LSTM~\cite{hochreiter1997lstm} cells (Sec.~\ref{subsec:scr-rnn}). 
To generate rich and fine-grained descriptions and reinforce the coarse-to-fine procedure of description generation, we adopt an auxiliary supervision, \textit{Linguistic Attributes}, hierarchically upon the outputs of the sequential LSTM cells, as shown in Fig.~\ref{fig:pipeline}. 
The proposed model is trained to minimize both the sentence loss as well as the binary cross-entropy losses (attribute losses) for caption generation.

\begin{figure*}[t]
\centering
\includegraphics[width=0.9\linewidth]{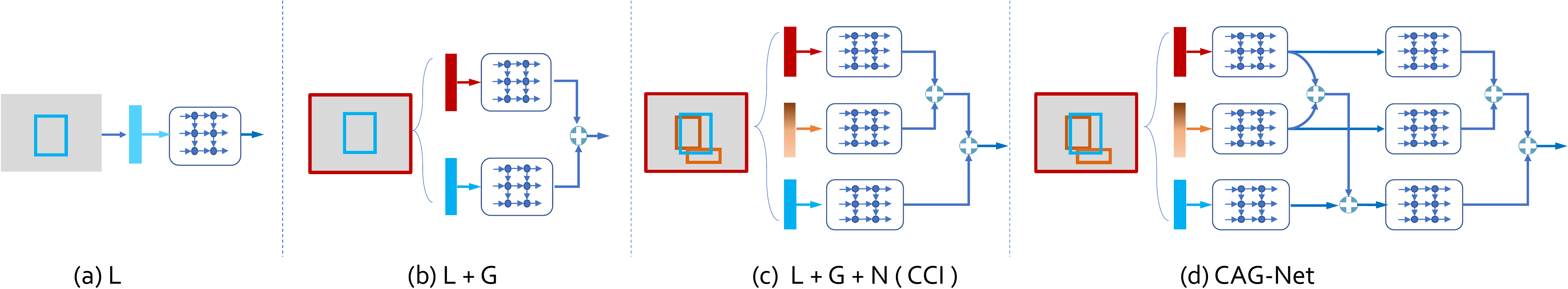}
\caption{Comparisons between different network structures. (a) L generates the descriptions separately after region proposals; (b) L + G generates descriptions with not only the {\color{cyan}local} feature but also the {\color{red}global} feature of the image; (c) L + G + N (CCI) integrates {\color{red}global}, {\color{orange}neighboring} and {\color{cyan}local} information for the target to generate descriptions; (d) CAG-Net by multiple LSTM cells is a stacked version of (c) CCI but supervised with hierarchical linguistic attribute losses.
}
\label{fig:module}
\end{figure*}

\begin{figure}[t]
\centering
\includegraphics[width=0.90\linewidth]{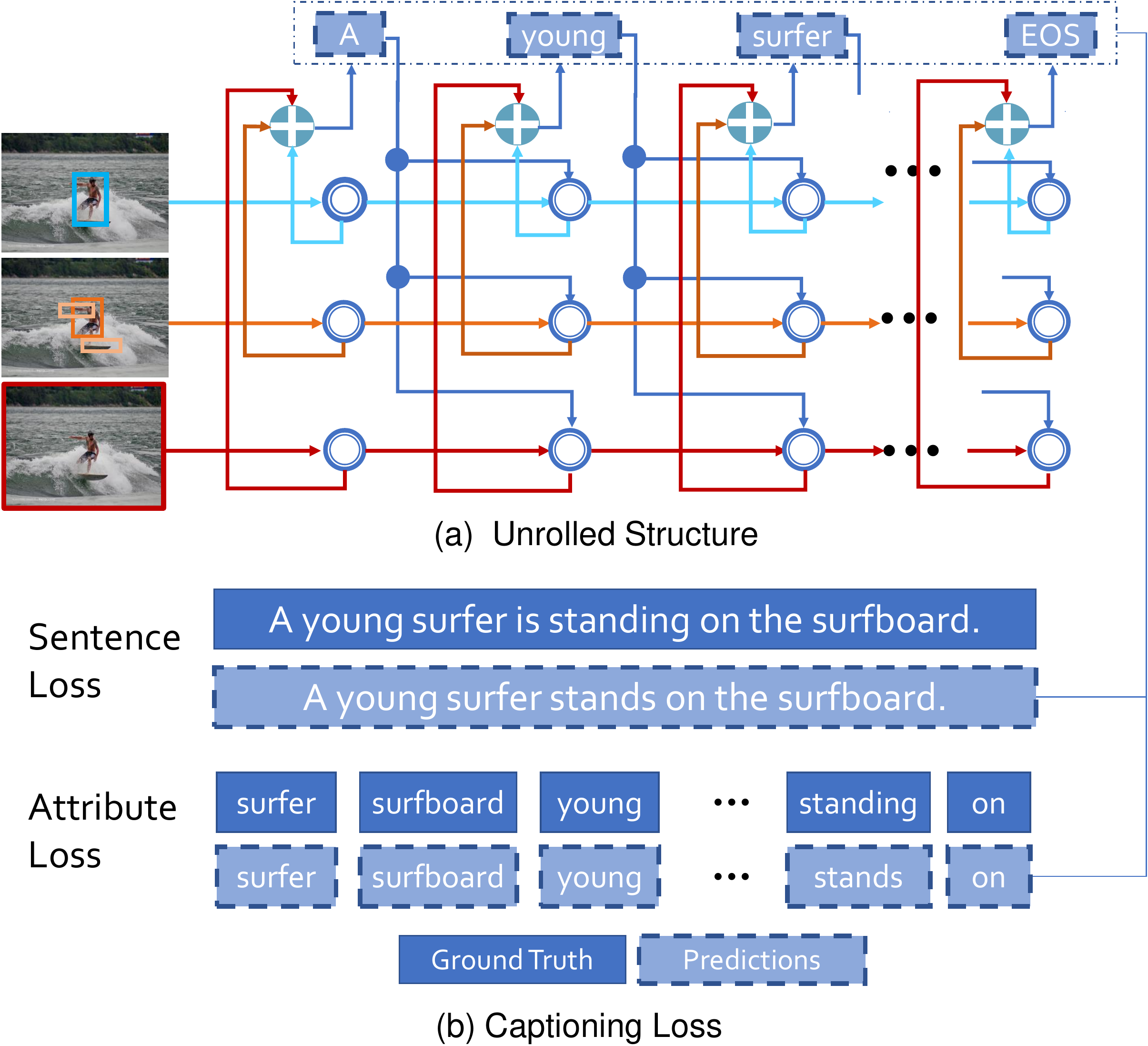}
\caption{
The unrolled structure of Contextual Cue Integrator (CCI). (a) Unrolled structure integrates the {\color{cyan}local} (in  blue) information and multi-scale context cues, \ie, {\color{red}global} (in red) and {\color{orange}neighboring} (in orange). The hollow circle stands for the LSTM cell while the plus sign for the feature fusion briefly. (b) The captioning loss consists of a sentence loss and an attribute loss. 
}
\label{fig:unrolled_rnn}
\end{figure}

\subsection{Contextual Feature Extractor}
\label{subsec:context}

Denote the regions-of-interest (ROIs) in an image as $\mathcal{R} = \{ \mathbf{R}_i | i=1,2,...,N \}$ and the entire image as $\mathbf{R}^*$.
The contextual features for the local region $\mathbf{R}_i$ are exploited from the multi-scale contextual cues of \textit{local} region $\mathbf{R}_i$, \textit{neighboring} region $\mathcal{R}_i^{n}=\mathcal{R}/\mathbf{R}_i$, and the \textit{global} region $\mathbf{R}^*$. 
For the target region $\mathbf{R}_i$, denote the \textit{local, neighboring} and \textit{global} features as $\mathbf{F}^l_{i}$, $\mathbf{F}^n_{i}$, and $\mathbf{F}^g_{i}$, respectively, where $\mathbf{F}^g_{i}$ refers to the features extracted from the entire input image and $\mathbf{F}^l_{i}$ is the feature of the target instance.
The \textit{Contextual Feature Extractor} (CFE) focuses on exploring the neighboring features $\mathbf{F}^n_{i}$ which can be formulated as $\mathbf{F}^n_{i}= f(\mathbf{R}_i, \mathcal{R}^n_{i})$.

We design a similarity graph based on region-level (\ie,~ROI-level) for neighboring ROIs aggregation, inspired by pixel-level non-local operations.
Non-local means~\cite{buades2005nonlocal} has been often used as a filter by computing a weighted mean of all pixels in an image, which allows pixels to contribute to the filtered response based on the patch appearance similarity.
Similarly, neighboring ROIs with similar semantic appearance are supposed to contribute more on the feature extraction for the target local instance.
Following the operation in~\cite{buades2005nonlocal}, we rewrite the formulation of $f(\mathbf{R}_i, \mathcal{R}^n_{i})$ as
\begin{equation}
f(\mathbf{R}_i, \mathcal{R}^n_{i}) = \sum_{\forall{j},j\neq i} {\mathcal{G}(\mathbf{F}_i^l, \mathbf{F}_j^l)\mathbf{F}^l_{j}},
\label{eq:non-local1}
\end{equation}
where $\mathcal{G}(\mathbf{F}_i^l, \mathbf{F}_j^l)$ is the appearance similarity between region $\mathbf{R}_i$ and $\mathbf{R}_j$, and $\mathbf{F}^l_i$ is the fixed-length local feature of region $\mathbf{R}_i$. The similarity $\mathcal{G}$ is the normalized cross correlation based on Gaussian function, formulated as,
\begin{equation}
\mathcal{G}(\mathbf{F}_i^l, \mathbf{F}_j^l) = \frac{exp ({\mathbf{F}^l_i}^{\top} \mathbf{F}^l_j)}{\sum_{\forall j,  j \neq i} exp ({\mathbf{F}^l_i}^{\top} \mathbf{F}^l_j )},
\label{eq:non-local2}
\end{equation}
where ${\mathbf{F}^l_i}^{\top} \mathbf{F}^l_j $ is dot-product similarity of cross correlation. Therefore, we can obtain the similarity graph for each target ROI with its neighboring ROIs in the image.

General object detection algorithm usually generates redundant region candidates (ROIs) to ensure the accuracy and robustness in region localization and detection.
However, in this case, the integrated \textit{neighboring} feature $\mathbf{F}^n_i$ will be contaminated by distant and independent proposals, and the amount of ROIs in $\mathcal{R}^n_{i}$ also tremendously increase the computation cost and noises in the environment.
Therefore, we sample a subset of $\mathcal{R}^n_{i}$ based on their spatial nearness such that the closer ROIs are more relative to the target ROI.
We sort ROIs in $\mathcal{R}^n_i$ based on the IoU (intersection over union) metric with the target region $\mathbf{R}_i$.
By sampling the top-$k$ proposals as $\hat{\mathcal{R}}^n_{i}$, the calculation of the \textit{neighboring} features can be accelerated as $\mathbf{F}^n_{i} = f(\mathbf{R}_i, \hat{\mathcal{R}}^n_{i})$.

\subsection{Attribute Grounded Caption Generator}
\label{subsec:scr-rnn}

In this section, we present a novel caption generator with two parts: (1) a \textit{Contextual Cue Integrator} to fuse contextual features produced by the CFE in Sec.~\ref{subsec:context}, and (2) an \textit{Attribute Grounded Coarse-to-Fine Generator} with coarse-level and fined-level linguistic attribute losses as the additional supervision to enhance the discriminativeness of the generated captions.

\noindent\textbf{Contextual Cue Integrator (CCI) -}
\label{subsubsec:multi_level_fusion}
The contextual cue integrator adopts multiple LSTM cells to hierarchically integrate the multi-scale contextual features into the localized features.
The \textit{local, neighboring} and \textit{global} features are spread through in the LSTM cells so as to generate context-aware descriptions for the target ROI.
These descriptions are fused together for the final captioning of the target region at each time step of LSTM.
The unrolled contextual cue integration module can be visualized in Fig.~\ref{fig:unrolled_rnn}(a).
The \textit{local} branch can be regarded as the backbone for the target while the \textit{global} and \textit{neighboring} branches are grouped as multi-scale contextual cues to provide complementary guidances.
Therefore, the contextual cues are adaptively combined at first, and they are then added to the local branch via a second adaptive fusion, as shown in Fig.~\ref{fig:module}(c).
The  importance of different levels' features is regularized by the adaptive weights, which are optimized during training the framework.

\begin{figure*}[t]
\centering
\includegraphics[width=0.9\linewidth]{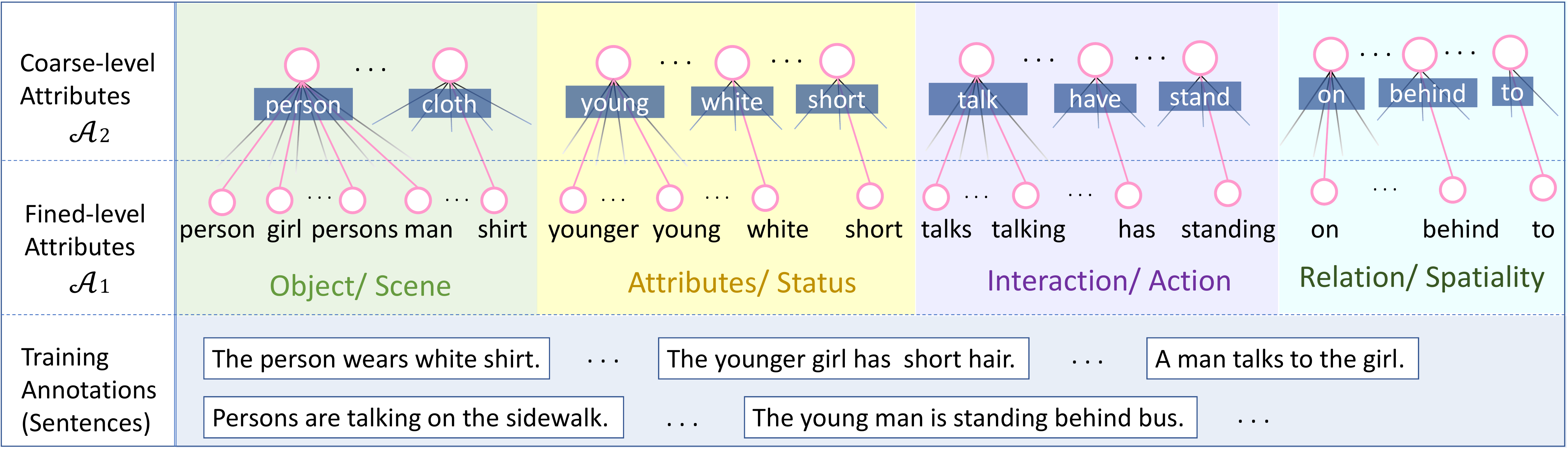}
\caption{Illustration of sentence itemization. Fined-level attributes $\mathcal{A}_1$: the original sentences of training annotations (bottom) are itemized to individual words and divided into four groups: object/scene (noun), attribute/status (adjective), interaction/action (verb) and relation/spatially (preposition). Coarse-level attributes $\mathcal{A}_2$: the individual words are normalized and clustered by semantical similarity for high-level words, \eg,~the \textit{girl} and \textit{man} in $\mathcal{A}_1$ belong to \textit{person} in $\mathcal{A}_2$.
}
\label{fig:sentence_loss}
\end{figure*}

\noindent\textbf{Attribute Grounded Coarse-to-Fine Generator -}
\label{subsubsec:attr_loss}
%
It is challenging in generating rich and accurate descriptions just by the sequential LSTMs. To this end, we increase its representative power by introducing a coarse-to-fine caption generation procedure with sequential LSTM cells, \ie, coarse stage and refined stage supervised with the auxiliary hierarchical linguistic attribute losses.

The linguistic attribute losses serve as the intermediate and auxiliary supervisions from coarse to fine in addition to the general sentence loss of captioning, implemented at each stage as shown in Fig.~\ref{fig:pipeline}.
The attribute losses are formatted as binary classification (\ie, exist or not) losses for each attribute separately during the training procedure. 
As shown in Fig.~\ref{fig:unrolled_rnn} (b), the attributes, \eg, \textit{surfer, standing, young} and \emph{on}, will be measured individually regardless of the speech order, similar as the multi-label classification for attribute recognition of objects. 

The subsequent LSTM layer (\texttt{refined} stage) is supposed to serve as the fine-grained decoders for the coarse regional descriptions generated by the preceding one (\texttt{coarse} stage).
The hidden vectors of LSTM cells produced by the coarse stage are taken as the disambiguating cues to the refined stage.
The outputs of \textit{global} and \textit{neighboring} branches at the coarse stage are used as the inputs of the respective branches directly at the refined stage.
The adaptive fusion of these three branches at the coarse stage is fed as the input at the refined stage. 
Meanwhile, these vectors are used for coarse-level attribute prediction.
The connection of the branches at the multiple stages is shown in the Fig.~\ref{fig:module}(d).
The final outputs of the word decoder at the refined stage are the generated descriptions for the target region. Meanwhile, these outputs are used for the fined-level attribute prediction as well.

These linguistic attributes are predicted from the outputs of the LSTMs during the training procedure and the unsolved problem here is how to get the ground-truth linguistic attributes.
In our work, the hierarchical \textit{linguistic attributes} are obtained by itemizing the sentences in the training split with natural language processing toolkit (NLTK).

\noindent 1) Fined-level attributes $\mathcal{A}_1$ for \texttt{refined} stage.
We distill the linguistic knowledge from the training annotations (sentences or phrases) to individual keywords/attributes, by the speech toolkit from NLTK
, as shown in Fig.~\ref{fig:sentence_loss}.
The reference sentences are parsed into four groups by the part-of-speech, \ie, nouns, adjectives, verbs and prepositions from the following aspects respectively: (1) The noun words are usually the labels of objects or scenes, \eg,~\textit{person}, \textit{bus}, \textit{sidewalk} and \textit{etc.}; (2) adjectives represent attributes or status, \eg,~\textit{young}, \textit{black}; (3) verbs are meanings of actions or interactions, \eg,~\textit{standing}, \textit{talks}; (4) prepositions for relations or spatiality, \eg,~\textit{behind}.
The fined-level attributes like \textit{surfer} and \textit{standing} are used
at the latter stage for the exact information extraction.

\noindent 2) Coarse-level attributes $\mathcal{A}_2$ for \texttt{coarse} stage.
We use the high-level semantically clustered attributes, \eg,~\textit{person}, \textit{stand} to stand for the major information.
We observe that labels with the same concept may have different singular and plural forms or different participles,~\eg,~\textit{persons} versus~\textit{person}, \textit{talks} versus \textit{talking}.
These words are normalized to a unified format by NLTK Lemmatizer,~\eg,~\textit{talk} from~\textit{talks} and \textit{talking}.
Furthermore, labels having closer semantic correlation (\eg,~\textit{girl} and \textit{man} are hyponyms of \textit{person}) need to be distinguished from other semantic concepts like \textit{cloth}, as shown in the top panel of Fig.~\ref{fig:sentence_loss}.
Therefore, we cluster the labels with their semantical similarities computed by Leacock-Chodorow distance~\cite{seco2004similarity}.
We find a threshold of $0.85$ is well-suited for splitting semantic concepts.
The coarse-level items like \textit{person} and \textit{stand} are used at the preceding stage for the key information extraction.

\section{Experiments}
\label{sec:experiments}

\subsection{Experiment Settings}
\label{subsec:exp_set}

\noindent \textbf{Dataset.}
Visual Genome (VG) region captioning dataset~\cite{krishna2017vg} is used as the evaluation benchmark in our experiments. For fair comparisons, we use the dataset of version 1.0 and the same train/validation/test splits as in~\cite{johnson2016densecap}, \ie, 77398 images for training and 5000 images each assigned for validation and test.

\vspace{0.15cm}
\noindent \textbf{Evaluation Metric.}
Following~\cite{johnson2016densecap}, the mean Average Precision (mAP) are measured across a range of thresholds for both accurate localization and language description jointly, inspired by the evaluation metrics in object detection~\cite{everingham2010pascal,lin2014microsoft} and image captioning~\cite{banerjee2005meteor}.
For localization, intersection over union (IoU) thresholds $.3$, $.4$, $.5$, $.6$, $.7$ are used while METEOR~\cite{banerjee2005meteor} score thresholds $0$, $.05$, $.1$, $.15$, $.2$, $.25$ used for language similarity. 
The average precision is measured across all pairwise settings, \ie, \textit{(IoU, METEOR)}, of these methods and report the mean AP (mAP),
which means that the mAP is computed for different IoU thresholds for localization accuracy, and different METEOR score thresholds for language similarity, then averaged to produce the final score.

To isolate the accuracy of language in the predicted captions without localization, the predicted captions are evaluated without taking into account their spatial positions.
Following~\cite{johnson2016densecap}, the references of each prediction are generated by merging ground truth across each image into a bag of reference sentences. 
 Apart from the mAP score introduced above, the METEOR score will be reported as the auxiliary evaluation metric, 
denoted as \textit{METEOR}.
%
Note that the references from all regions in an image only offer the global and coarse ground truth descriptions. 

\vspace{0.10cm}
\noindent \textbf{Implementation Details.}
We use VGG-16~\cite{krishna2017vg} pretrained on ImageNet~\cite{deng2009imagenet} as the network backbone.
As shown in Fig.\ref{fig:pipeline}, we use $6$ LSTM cells in total, \ie, one LSTM for \textit{local, neighboring, global} features respectively at each stage.
The newly-introduced layers and LSTM cells are randomly initialized and our proposed CAG-Net is optimized by end-to-end training.
The implementations are based on Faster RCNN~\cite{ren2015faster} using Caffe~\cite{jia2014caffe}, and the networks are optimized via stochastic gradient descent (SGD) with base learning rate as $0.001$.
The input image is re-sized to have a longer side of 720 pixels and 256 proposals are sampled per image in each forward pass of training.
The LSTM cell for sequential modeling is used with $512$ hidden nodes. The most $10,000$ frequent words in the training annotations are remained as the vocabulary and other words are collapsed into a special $<$\texttt{UNK}$>$ token under the same conditions as in~\cite{yang2017densecap}. Following~\cite{johnson2016densecap}, we discard all sentences with more than 10 words ($7\%$ of annotations), that is the time length of the LSTMs is $10$.

The complete framework are training by end-to-end in our experiments.
The losses of our framework are from two aspects: 
1) Location: Smooth $\ell_1$ loss for bounding box regression ({$\mathcal{L}_{bbox}$) and softmax loss for binary foreground/background classifier ({$\mathcal{L}_{cls}$),
2) Caption: Cross entropy loss of sentences for description generation ({$\mathcal{L}_{sent}$), following~\cite{yang2017densecap} and binary cross entropy loss for linguistic attribute recognition ({$\mathcal{L}_{attr}$) . 
The total loss function is formulated as $\mathcal{L} = \mathcal{L}_{sent} + \alpha\mathcal{L}_{bbox} + \beta\mathcal{L}_{cls} + \gamma\mathcal{L}_{attr}$, where $\alpha=0.1$, $\beta=0.1$ and $\gamma=0.01$ in our experiments with the empirical values.

In evaluation, we follow the settings of~\cite{johnson2016densecap} for fair comparisons. $300$ proposals with the highest predicted confidence are remained after non-maximum suppression (NMS) with IoU threshold $0.7$.
We use efficient beam-1 search to produce region descriptions, where the word with the highest confidence is selected at each time step.
With another round NMS with IoU threshold $0.3$, the remaining regions and their generated descriptions are used for the final evaluation. To establish an upper bound regardless of region proposals, we evaluate the models on ground truth bounding boxes as well, marked as GT in the tables.

\begin{table}[t]
\centering

\footnotesize
\begin{tabular}{M{2.0cm}||M{0.75cm} M{1.25cm} | M{0.75cm} M{1.25cm}}
\hlinew{1.1pt}
 \multirow{2}{*}{Methods} &\multicolumn{2}{c|}{RPN}&\multicolumn{2}{c}{GT}\\
  & mAP  & METEOR & mAP  & METEOR \\
\hline
CAG-Net & \textbf{10.51} & \textbf{0.279} & \textbf{36.29} & \textbf{0.316}  \\
\hline
T-LSTM~\cite{yang2017densecap} & 9.31 & 0.275 & 33.58 & 0.307 \\
FCLN~\cite{johnson2016densecap} & 5.39 & 0.273 & 27.03 & 0.305 \\
\hlinew{1.1pt}
\end{tabular}

\caption{Quantitative results on Visual Genome comparing with state-of-the-art methods, T-LSTM~\cite{yang2017densecap} and FLCN~\cite{johnson2016densecap}.  Results in bold indicate the best performance. The metrics on T-LSTM, \ie, METEOR, are not provided in the paper and we measure these metrics using the model provided by the authors.}
\label{tb:state_of_the_arts}
\end{table}

\subsection{Comparison with State-of-the-Art Methods}
\label{subsec:comparison_methods}

We quantitatively compare the performances of the proposed \textit{Context and Attribute Grounding Dense Captioning} (CAG-Net) model with the previous state-of-the-arts, \ie, FCLN\cite{johnson2016densecap} and T-LSTM~\cite{yang2017densecap}.
FCLN~\cite{johnson2016densecap} introduces a fully differentiable layer for dense localization. The captioning per region is generated solitarily without any message passing from the contextual features. T-LSTM~\cite{yang2017densecap} designs network structures that incorporate two novel parts: joint inference for accurate localization and context fusion with the global scene for accurate description regardless of the interactions among the relative regions.

The comparison experiments use the same settings as the prior arts, shown in Tab.~\ref{tb:state_of_the_arts}. The CAG-Net significantly outperforms these methods by achieving a gain on mAP score from $9.31\%$ to $10.51\%$ using RPN and from $33.58\%$ to $36.29\%$ using the ground truth bounding boxes compared to the previous state-of-the-art, T-LSTM~\cite{yang2017densecap}.
The performance gains are mainly from the benefits of attribute grounded coarse-to-fine description generation using the contextual feature extractor and message integration among the regions.
The proposed CAG-Net presents a strong capability in capturing the correlation among relative regions and generating more accurate descriptions.

It is observed that the METEOR scores of different methods are approximate while the mAP scores have a large margin. 
That is because that the METEOR score for the predicted caption is calculated by using all ground truth descriptions of all the regions in the image as the references. These references are coarse and may not be accurate for a certain region. 
In the following ablation study (Sec.~\ref{sec:ablation}), we mainly focus on the comparison of mAP scores.

\begin{figure*}[t]
\centering
\includegraphics[width=0.95\linewidth]{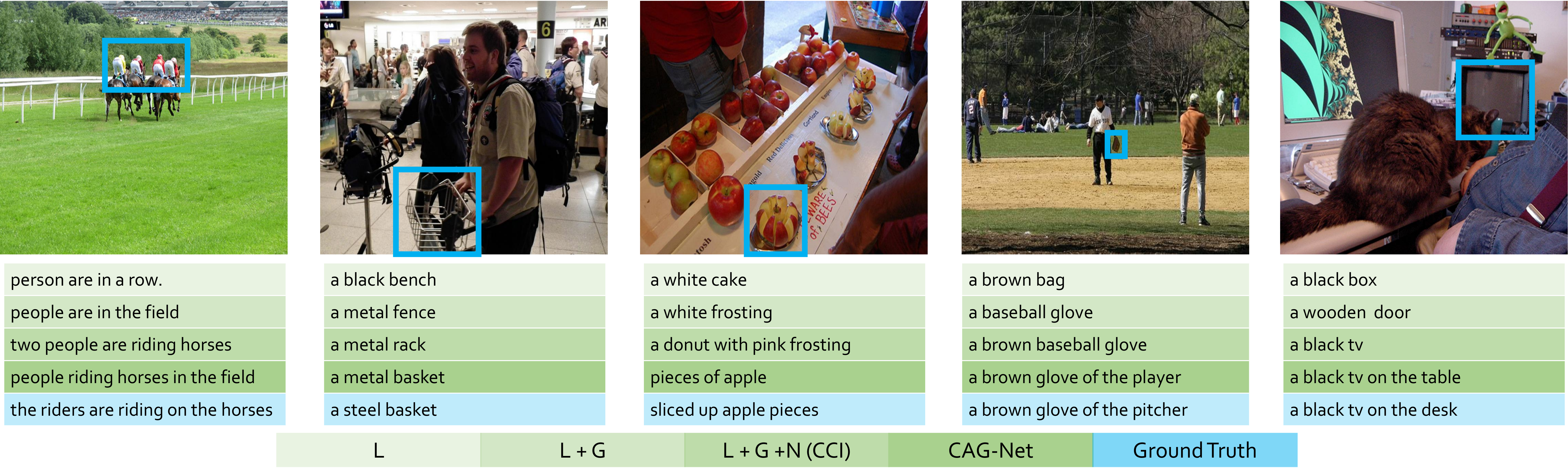}
\caption{Qualitative results of CAG-Net compared with variants of different module configurations on VG dataset, \ie, (a) L(Local Cue), (b) L+G (Local and Global Integration), (c) L+G+N (CCI) (Local, Global and Neighboring Integration). 
}
\label{fig:results}
\end{figure*}

\begin{table}[t]
\centering
\footnotesize
\vspace{0.1cm}
\begin{tabular}{M{0.75cm} M{0.75cm}|M{1.5cm} M{0.95cm} M{0.85cm} M{0.7cm}}
\hlinew{1.1pt}
\multicolumn{2}{c|}{Methods} & CAG-Net & L+G+N & L+G & L \\
\hline
\multirow{2}{*}{mAP} & RPN & \textbf{10.51} & 9.55 & 7.97 & 6.31 \\
& GT & \textbf{36.29} & 33.50 & 31.77 & 26.70\\
\hlinew{1.1pt}

\end{tabular}
\caption{Ablation study on CAG-Net compared the variants of the contextual cue integration module, \ie,  1) \texttt{L}, local cue without neighboring nor global features, 
2) \texttt{L+G}, local and global cue integration and 3) \texttt{L+G+N}, local, global and neighboring integration  without stacking contextual cue integration modules.
Results in bold indicate the best performance.
}
\label{tb:ablation_snlrnn}

\end{table}

\section{Ablation Study}
\label{sec:ablation}

\subsection{CAG-Net}
\label{subsec:ablation_snlrnn}

\noindent\textbf{Attribute Grounded Caption Generator with Contextual Cues.}
To demonstrate the benefits of multi-scale contexts and attribute grounded captioning module, we compare the results of CAG-Net in Fig.~\ref{fig:module} (d) with the variants by removing individual cue step by step, \ie, 1) \texttt{L}, local cue as the baseline without either contextual neighboring or global features as shown in Fig.~\ref{fig:module} (a), 
2) \texttt{L+G}, local and global cue integration without contextual neighboring cues in Fig.~\ref{fig:module} (b) and 3) \texttt{L+G+N}, local, global and neighboring integration without stacking contextual cue integration modules in Fig.~\ref{fig:module} (c), defined as CCI in Sec~\ref{subsubsec:multi_level_fusion}.
 The quantitative results are reported in Tab.~\ref{tb:ablation_snlrnn}.

Compare with basic \texttt{L}, the mAP of \texttt{L+G+N} can be improved from $6.31\%$ to $9.55\%$ using RPN and from $26.70\%$ to $33.50\%$ using ground truth boxes by involving contextual feature extractor and message integration while the mAP of \texttt{L+G} which only includes the global cues achieves $7.97\%$ using RPN and $31.77\%$ using ground truth bounding boxes.
The significant improvement demonstrates the importances of contextual cue integration between multi-scale contexts and individual regions for region generation and the contextual cues, \ie, \textit{global} and \textit{neighboring} make a certain contribution to improving the final performances.
Furthermore, with the assistance of the linguistic attribute losses, the mAP of CAG-Net achieves $10.51\%$ in mAP using RPN by a gain of $0.96\%$ compared to \texttt{L+G+N} (CCI) while a gain of $1.79\%$ using the ground truth bounding boxes.
No doubt that the generated descriptions are more accurate and rich for the regions when adopting attribute grounded coarse-to-fine captioning module.

The qualitative results are shown in Fig.~\ref{fig:results}. 
The descriptions directly generated by the target regions are fallible due to lack of enough visual information, \eg, mistaking the \textit{baseball glove} for \textit{a brown bag}, the \textit{apple pieces} for \textit{a white cake} and the \textit{steel basket} for \textit{black bench}.
The involved global cues of the image also lead to deviation, \eg, the \textit{tv in the room} is mistakenly predicted as \textit{a wooden door}, although positive effect sometimes, \eg, the \textit{glove} is accurately predicted with the assistance of the global image feature. Furthermore, the coarse-to-fine generation module  will reinforce more rich descriptions \textit{a black tv on the table} compared with individual module \textit{a black tv} shown in the figure.
The results shows the excellent performance of the proposed context and attribute grounded generation structure for dense captioning.

\begin{table}[t]
\centering
\footnotesize
\begin{tabular}{M{0.35cm} M{0.45cm}|M{0.7cm} M{0.75cm} M{0.75cm} M{0.75cm} M{0.65cm} | M{0.5cm} }
\hlinew{1.1pt}
& &\multicolumn{5}{c|}{CAG-Net} & \\
\multicolumn{2}{c|}{Methods} & $(\mathcal{A}_2,\mathcal{A}_1)$  & $(\mathcal{A}_1, \mathcal{A}_1)$  & $(\mathcal{A}_2, \mathcal{A}_2)$ & $(1k, 1k)$ & $(-,-)$ & CCI\\
\hline
\multirow{2}{*}{mAP} & RPN & \textbf{10.51} & 9.93 & 9.99 & 9.95 & 9.59 & 9.55\\
& GT & \textbf{36.29} & 34.98 & 35.17 & 35.02 & 33.78 & 33.50\\

\hlinew{1.1pt}
\end{tabular}
\caption{Ablation study on CAG-Net compared the variants of linguistic attribute losses, \ie,
1) $(\mathcal{A}_2 , \mathcal{A}_1)$, with the proposed coarse-to-fine attributes, 
2) $(\mathcal{A}_1, \mathcal{A}_1)$, only with the fined-level attributes $\mathcal{A}_1 $, 3) $(\mathcal{A}_2, \mathcal{A}_2)$, only with the coarse-level attributes $\mathcal{A}_2 $, 4) $(1k, 1k)$, replacing the proposed attributes with the top $1k$ attributes, 5) $(-,-)$, stacked structure without any attributes, 6) CCI, just one stage without attributes.
Results in bold indicate the best performance.
}
\label{tb:ablation_attrloss}
\end{table}

\vspace{0.10cm}
\noindent\textbf{Linguistic Attribute Losses.}
To demonstrate the benefits of the proposed linguistic attribute losses, we compare the performances of CAG-Net with the variants of linguistic attributes by 
1) ``$(-,-)$'', removing all the auxiliary linguistic attribute losses in the framework, 
2) ``$(\mathcal{A}_1, \mathcal{A}_1)$'', only with the fined-level attributes $\mathcal{A}_1 $ at two stages, 
3) ``$(\mathcal{A}_2, \mathcal{A}_2)$'', only with the coarse-level attributes $\mathcal{A}_2 $ at two stages, 
4) ``($1k$,$1k$)'', replacing the proposed linguistic attributes with the top $1k$ attributes (the top $1k$ most frequent words in the vocabulary) at two stages.

The results are shown in Tab.~\ref{tb:ablation_attrloss} and CAG-Net with the proposed coarse-to-fine linguistic attributes is denoted as  ``$(\mathcal{A}_2, \mathcal{A}_1)$''.
Compared with ``CCI'',  CAG-Net without any attributes (denoted as ``$(-,-)$'' in Tab.\ref{tb:ablation_attrloss}) gets the approximate results because the navie stacking description generation modules cannot significantly improve the performance although with more parameters. 
In contrast, the attribute grounded structure with the proposed coarse-to-fine attributes can achieve a gain from $9.59\%$ to $10.51\%$(using RPN) and from $33.78\%$ to $36.29\%$ (using ground truth boxes) because of the auxiliary hierarchical supervision of the proposed linguistic attribute losses.
Furthermore, to evaluate the effectiveness of the coarse-to-fine structure,  we compare CAG-Net, \ie, ``$(\mathcal{A}_2, \mathcal{A}_1)$'', with the variants of different linguistic attributes, \ie, `$(\mathcal{A}_1, \mathcal{A}_1)$'', ``$(\mathcal{A}_2, \mathcal{A}_2)$'' and ``($1k$, $1k$)''.
Without the coarse-to-fine strategy at two stages, the stacked structures with different attributes cannot achieve as good performance as CAG-Net both using RPN and using ground truth bounding boxes.
It is significant that the proposed linguistic attribute losses from the coarse to fine stage can improve the description generation of target regions.

\begin{table}[t]
\centering
\small
\begin{tabular}{M{0.35cm} M{0.5cm}|M{0.8cm} M{0.8cm} | M{0.65cm} M{0.65cm} M{0.65cm} M{0.65cm} }
\hlinew{1.1pt}
\multicolumn{2}{c|}{Methods} & Random & Nearest & SG & FC & AVE & MAX\\
\hline
\multirow{2}{*}{mAP} & RPN & 8.626 & 9.144 & 9.315 & 8.132  & 7.981 & 8.024 \\
& GT & 32.274 & 33.411 & 33.412 & 30.272 & 29.937 & 30.121 \\

\hlinew{1.1pt}
\end{tabular}
\caption{Results of Contextual Feature Extractor with different settings. ``Random'' means selecting the contextual neighboring regions randomly from all the regions in the image. ``Nearest'' means selecting the relative regions from the nearest ones sorted by the IoU scores.  ``SG'' means fusing these neighboring regions with similarity graph. ``FC'' means fusing $k$-sorted neighboring regions with fully connected layer. ``AVE'' means average-pooling of $k$-sorted neighboring regions. ``MAX'' means max-pooling of $k$-sorted neighboring regions.
}
\label{tb:non_local_spatial}
\end{table}

\subsection{Contextual Feature Extractor}
\label{subsec:ablation_nlrnn}
In this section, we compare the performances of Contextual Feature Extractor (CFE) with a set of variants achieved by changing one of the hyper-parameters or settings step by step to explore the best practice of the proposed contextual feature extractor. As for the generation structure, we use CCI instead of CAG-Net due to the faster speed and less computation cost.

\vspace{0.10cm}
\noindent\textbf{Contextual Feature Extractor of $k$-nearest neighboring regions performs best.}
To explore the benefits of similarity graph in Contextual Feature Extractor in our framework, we replace the similarity graph in the CCI shown in Fig.~\ref{fig:module} (c) with 1) ``FC'', the fully-connected layer, 2) ``MAX'', max-pooling layer, 3) ``AVE'', average-pooling layer after concatenating all the feature vectors of $k$ neighboring regions. 
The results are shown in Tab.~\ref{tb:non_local_spatial}. 
The similarity graph operation can improve all the evaluation metrics compared with the simple fully-connected/ max-pooling/ average-pooling operation after concatenating all the feature vectors of $k$ neighboring regions.
That's because the similarity graph not only includes the visual features of $k$ neighboring regions but also utilizes the relation between the target region and the neighboring region.  
Furthermore, Tab.~\ref{tb:non_local_spatial} shows that the nearest-neighbor regions (``Nearest'') perform better than the regions randomly-selected from all the regions in the image (``Random'') due to more correlated regions involved in the description generation.

\vspace{0.10cm}
\noindent\textbf{Contextual Feature Extractor with hyper-parameter $k=20$ outperforms others.}
The number of neighboring regions is worth investigating because it can be used to find a trade-off between the effective message passing and the noises from non-correlated proposals in the image.
We validate the number of neighboring regions among $10, 20, 30, 50$ and $100$ of CCI. The results are reported in Tab.~\ref{tb:num_k}. 
We adopt $k$ as $20$ for further experiments for the best performance ($9.144\%$) on mAP considering region localization and description jointly.

\begin{table}[t]
\centering
\small
\begin{tabular}{M{0.6cm} M{0.6cm}|M{0.75cm} M{0.75cm} M{0.75cm} M{0.75cm} M{0.75cm} }
\hlinew{1.1pt}
\multicolumn{2}{c|}{$k$} & 10 & 20 & 30 & 50 & 100 \\
\hline
\multirow{2}{*}{mAP} & RPN & 8.915 & 9.144 & 9.109 & 8.804 & 8.749 \\
& GT & 33.260 &33.412 & 33.411 & 33.389 & 33.089 \\

\hlinew{1.1pt}
\end{tabular}
\caption{Results with different numbers of $k$-nearest regions for neighboring features in the Contextual Feature Extractor. 
The results are reported when hyper-parameter $k$ is set as $10, 20, 30, 50, 100$ respectively.
}
\label{tb:num_k}
\end{table}

\section{Conclusion}
\label{sec:conclusion}
In this paper, we propose a novel end-to-end framework for dense captioning, named as Context and Attribute Grounded Dense Captioning (CAG-Net) by utilizing the visual information of both the target region and multi-scale contextual cues, \ie, \emph{global} and \emph{neighboring}.
The proposed contextual feature extractor exploits the message passing between the target region and $k$-nearest neighboring regions in the image while the attribute grounded contextual cue integration modules reinforce rich and accurate description generation.
To enhance the description generation for the regions, we extract linguistic attributes from the reference sentences as the auxiliary supervision at each stage during the training process.
Extensive experiments demonstrate the respective effectiveness and significance of the proposed CAG-Net on the challenging large-scale VG dataset.

\vspace{0.3cm}
\noindent\textbf {Acknowledgment}
This work is supported in part by the National Natural Science Foundation of China (Grant No. 61371192), the Key Laboratory Foundation of the Chinese Academy of Sciences (CXJJ-17S044) and the Fundamental Research Funds for the Central Universities (WK2100330002, WK3480000005), in part
by SenseTime Group Limited, the General Research Fund sponsored by the Research Grants Council of Hong Kong (Nos. CUHK14213616, CUHK14206114, CUHK14205615, CUHK14203015, CUHK14239816, CUHK419412, CUHK14207-814, CUHK14208417, CUHK14202217), the Hong Kong Innovation and Technology Support Program (No.ITS/121/15FX).

{\small
\bibliographystyle{ieee_fullname}
\bibliography{ID1800.bbl}
}

\clearpage

\section{Appendix}

\begin{figure}[htb]
\centering
\includegraphics[width=0.99\columnwidth]{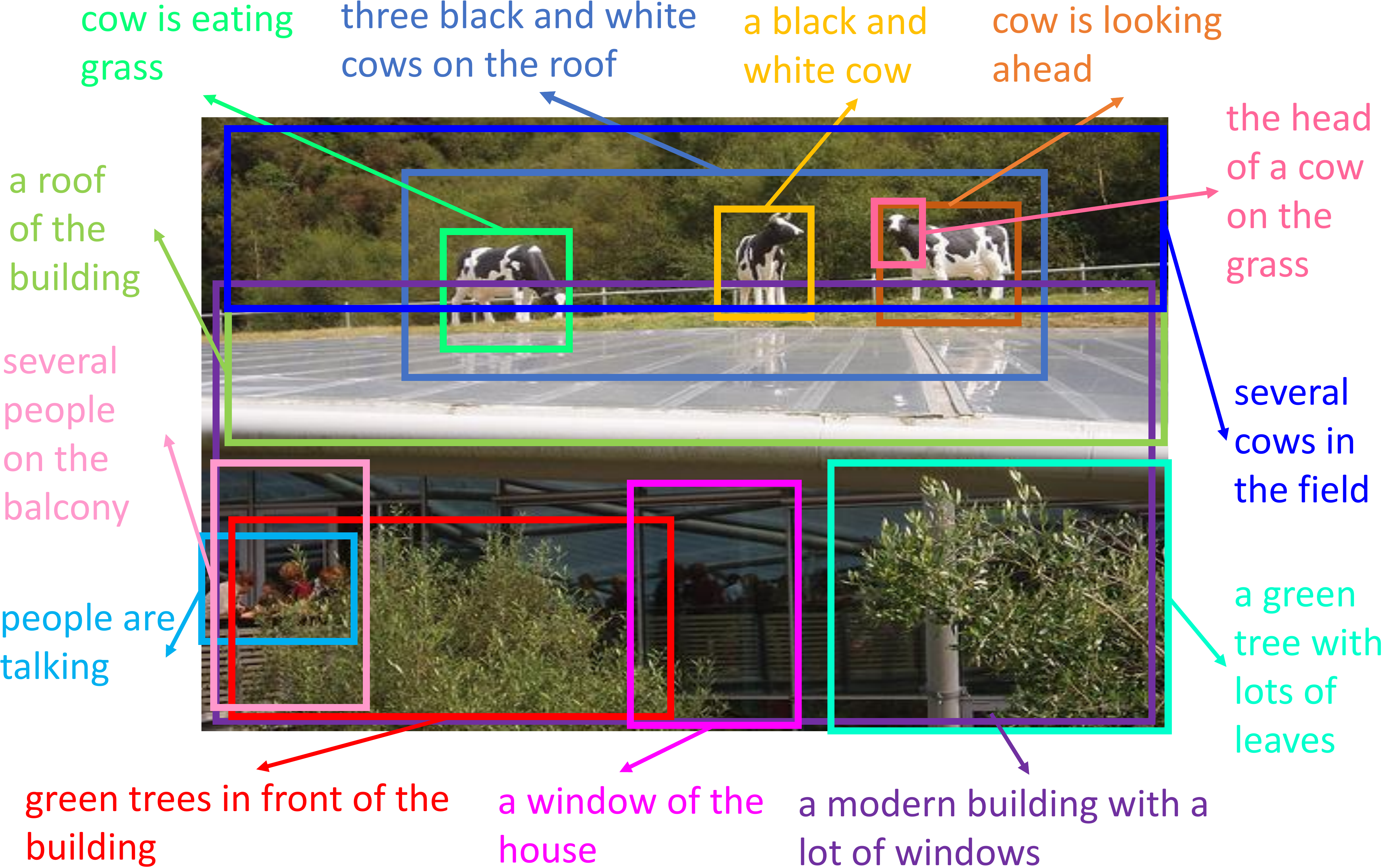} \\
\vspace{0.6cm}
\includegraphics[width=0.99\columnwidth]{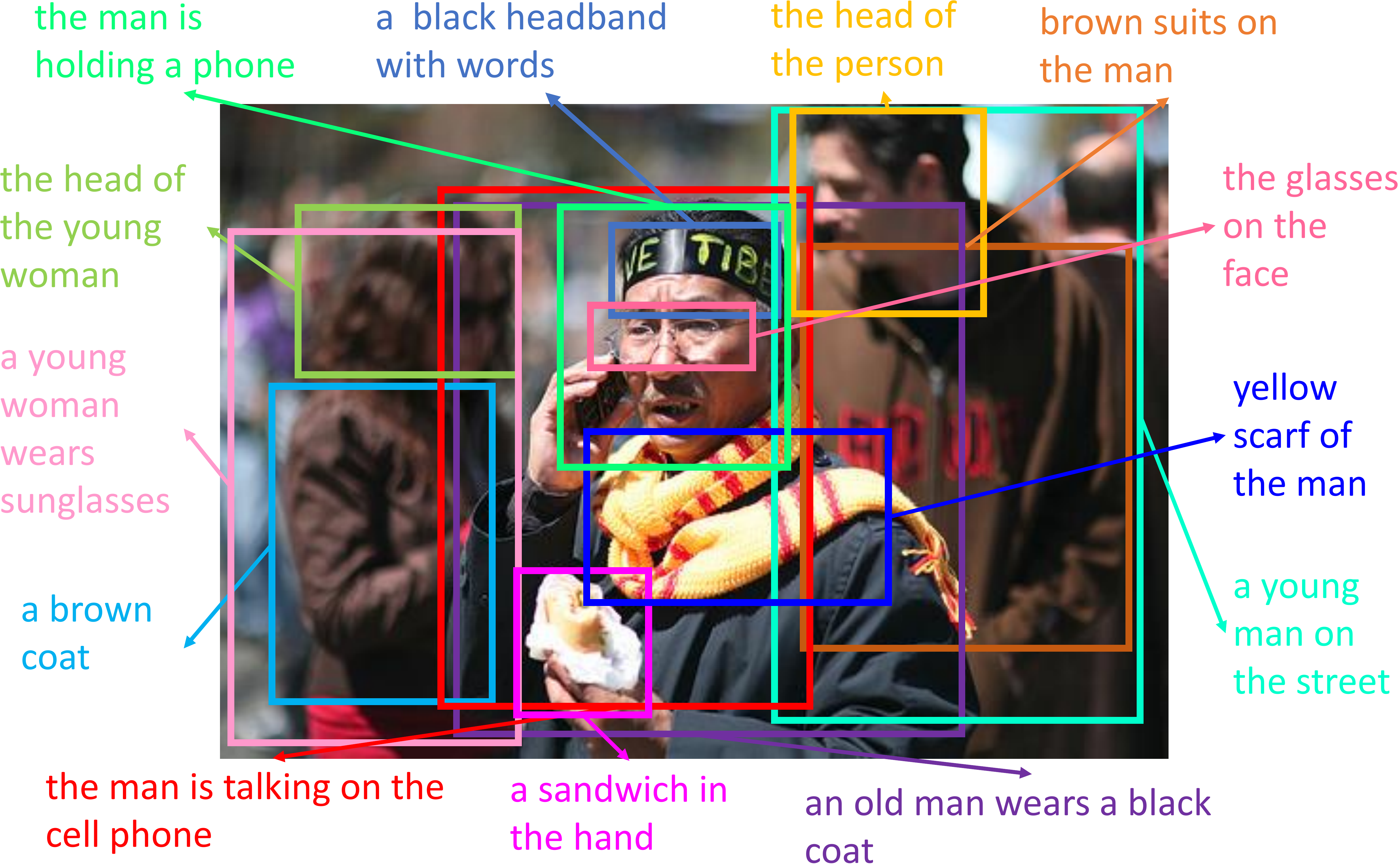} 
\caption{
Qualitative results on dense captioning by the proposed CAG-Net. The bounding box and description of the target region are in a consistent color. The images are overlaid with the most confident region descriptions. Best viewed in color}
\label{fig:densecap}
\end{figure}

\subsection{Dense Captioning Results of CAG-Net}

We present several instances of qualitative results on dense captioning by the proposed Context and Attribute Grounded Dense Captioning (CAG-Net) in Fig.~\ref{fig:densecap}.
The images are overlaid with the most confident region descriptions generated by our CAG-Net where the regions of interest are extracted by Region Proposal Network (RPN). The bounding box and description of the target region are in a consistent color.
Since the dense ROIs in the image overlap with each other frequently, their corresponding descriptions are obtained by the message passing and valid information sharing.
For example, it is not easy to describe the region (in light blue) in the top image of Fig.~\ref{fig:densecap} with ``three black and white cows on the roof'' only utilizing the appearance of the region while the cue of ``roof'' can be obtained by the neighboring regions in the image, \eg, the region of the ``building'' (in purple).
%

\begin{figure}[t]
\centering
\includegraphics[width=0.95\linewidth]{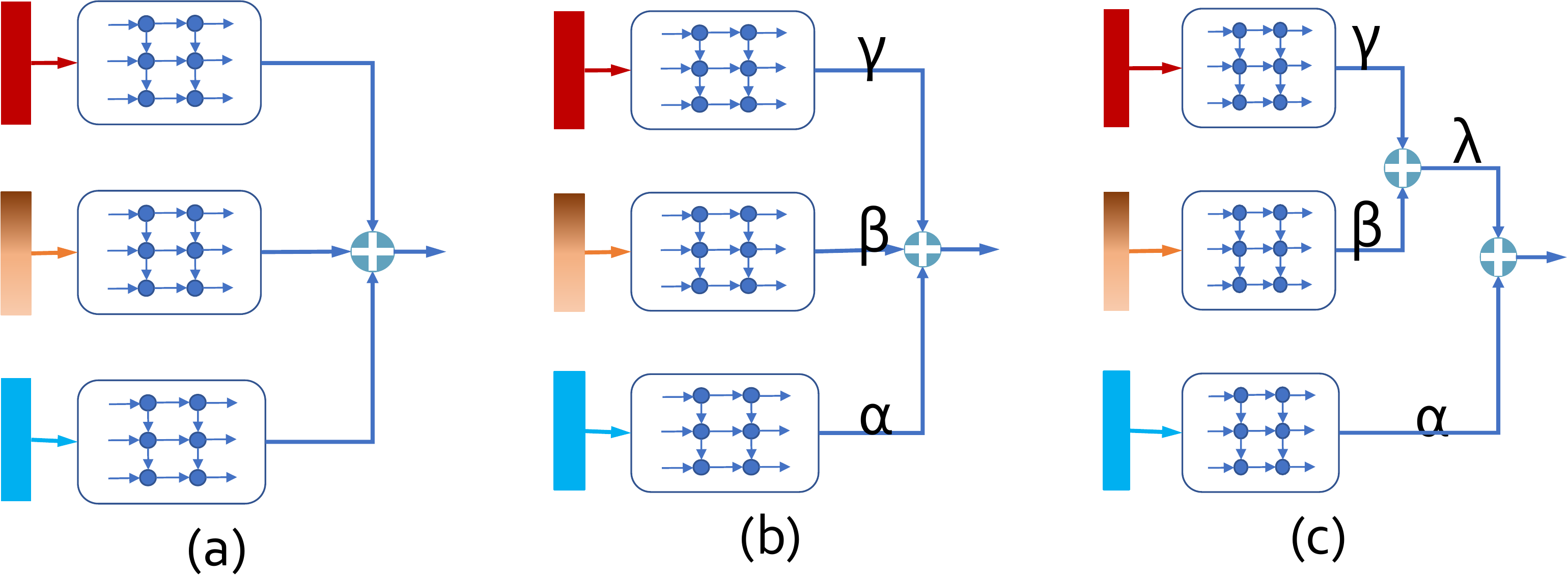}
\caption{ The illustration of fusion between the target (in blue) and multi-scale context cues, \ie global (in red) and neighboring (in orange). (a) Averaging fusion. (b,c) Adaptive fusion.}
\label{fig:fusion}
\end{figure}

\begin{table}[t]
\centering
\caption{Results of CCI with different integration types on Visual Genome dataset. The ``sum'', ``product'' and ``concat'' refer to the structures in Fig.~\ref{fig:fusion} (a) corresponding to the operations of sum, product and concatenation in channel dimension respectively. The ``adaptive1'' stands for the structure shown in Fig.~\ref{fig:fusion} (b) while ``adaptive2'' for Fig.~\ref{fig:fusion} (c).
}
\label{tb:fusion}
\small
\begin{tabular}{M{0.45cm} M{0.45cm}|M{0.65cm} M{0.85cm} M{0.85cm} | M{0.99cm} M{0.99cm} }
\hlinew{1.1pt}
\multicolumn{2}{c|}{$k$} & sum & product & concat & adaptive1 & adaptive2 \\
\hline
\multirow{2}{*}{mAP} & RPN & 9.239 & 8.721 & 9.279 & 9.326 & \textbf{9.474} \\
& GT & 32.754 & 32.719 & 32.731 & 33.589 & \textbf{33.603} \\

\hlinew{1.1pt}
\end{tabular}
\end{table}

\subsection{Ablation Study on Integration Types of CCI}

The \textit{local, neighboring} and \textit{global} features broadcast in the LSTM cells in parallel to generate descriptions with respect to their different spatial relationships with the target region.
These features are fused together for the final captioning of the target region at each time step. In this paper, we propose an \textit{Adaptive Fusion} strategy. The details can be found in Sec.3.2 (\emph{Contextual Cue Integrator (CCI)}) in the paper. In this supplementary material, we further provide a discussion on different types of integration strategies, \ie, \textit{averaging fusion} and \textit{adaptive fusion}, as shown in Fig.~\ref{fig:fusion}.

\vspace{0.10cm}
\noindent \textbf{Averaging Fusion.}
As shown in Fig.~\ref{fig:fusion} (a), the outputs of LSTM cells in three branches (for \emph{local}, \emph{neighboring} and \emph{global}) are merged with equal contributions, by some simple vector fusion operations, \eg, \texttt{sum}, \texttt{product} and \texttt{concatenation} in the channel dimension. 

\vspace{0.15cm}
\noindent \textbf{Adaptive Fusion.}
Instead of the simple averaging fusion operation, adaptive fusion of these multi-level descriptions are exploited in our framework.
The weights of these three kinds of features are learned during the end-to-end training of the entire framework, as shown in Fig.~\ref{fig:fusion} (b). 
Furthermore, in Fig.~\ref{fig:fusion} (c), the \textit{local} branch can be regarded as the backbone while the \textit{global} and \textit{neighboring} branches are grouped as multi-scale contextual cues to provide complementary guidances. 
Therefore, the contextual cues are combined in the first integration level and then added to the local branch with a second-level adaptive fusion.
The adaptive weights will be optimized during the training process.

\vspace{0.1cm}
\noindent\textbf{Experiment Results.} 
The comparison results with various integration types of CCI are reported in Tab.~\ref{tb:fusion}. 
For averaging fusion, the product operation performs best.
The hierarchical adaptive fusion structure shown in Fig.~\ref{fig:fusion} (c) outperforms the others and get the highest score of $9.474\%$ with RPN and $33.603\%$ using groundtruth bounding boxes. 
It shows that the hierarchical adaptive weights trained by end-to-end can get the best performance by adjusting the contributions of multi-scale contexts to the final descriptions of the targets.

\end{document}